# How far from automatically interpreting deep learning

Jinwei Zhao[a,b], Qizhou Wang[a,b], Yufei Wang[a,b], Xinhong Hei[a,b+], Yu Liu[a,b]

[a]School of Computer Science and Engineering, Xi'an University of Technology, Xi'an 710048, China

[b]Shaanxi Key Laboratory of Network Computing and Security Technology (Xi'an University of Technology), Xi'an 710048, China

+ Corresponding author: Xinhong Hei

E-mail: xinghonghei@xaut.edu.cn

Phone: +86-029-82312196(2016)

**Abstract**: In recent years, deep learning researchers have focused on how to find the interpretability behind deep learning models. However, today cognitive competence of human has not completely covered the deep learning model. In other words, there is a gap between the deep learning model and the cognitive mode. How to evaluate and shrink the cognitive gap is a very important issue. In this paper, the interpretability evaluation, the relationship between the generalization performance and the interpretability of the model and the method for improving the interpretability are concerned. A universal learning framework is put forward to solve the equilibrium problem between the two performances. The uniqueness of solution of the problem is proved and condition of unique solution is obtained. Probability upper bound of the sum of the two performances is analyzed.

## 1. Introduction

Safe, controllable and credible artificial intelligence has been the goal which the humanity has been pursuing. In the field of deep learning, in order to achieve this goal, it is needed for learning algorithm to really interact with the humanity and it is also indispensable for the learning algorithm to have the ability to correct errors, so as to avoid a prediction model with serious errors caused by unnecessary deviation in training data. So, it is necessary to establish a learning algorithm for capturing and learning causal relationships in the world around us. However, recently, all of this is out of reach. The reason is that the prediction model and its training process are not yet understood by human being. In other words, there is a gap between the deep learning model and the cognitive modes from human being.

For shrinking the gap, two general interpretation methods about deep learning model were identified by Lipton [1]: posting interpretation and transparent interpretation. For deep learning, the current mainstream methods are mainly from three aspects: hidden layer analysis method [2-4], simulation model method [5], attention mechanism[6-8].

We posits that how to make the prediction model and training process understood by us ascribe to an optimization problem which can promote the interpretability of the prediction model and make the model more suitable to its causality or discover faults in the causality.

## 2. Learning framework of traditional machine learning

Suppose $X$ is a compact domain or a manifold in Euclidean space and $Y \in R^k$, $k = 1$, $\rho$ is a Borel probability measure of a space $Z = X \times Y$. $f_\rho: X \to Y$ as $f_\rho(x) = \int_Y y \, d\rho(y|x)$ is defined. The function $f_\rho$ is a regression function of $\rho$. In machine learning, $\rho$ and $f_\rho$ are unknown. At some conditions, an edge probability measure $\rho_X$ of $X$ is known. The goal of the learning is to find the best approximation of $f_\rho$ in a functional space. Therefore, Tihonov regularization learning framework[9-10] can be obtained

$$f_{z,\gamma} = \mathrm{argmax}_{f \in \mathcal{H}_K} \left\{ \frac{1}{m} \sum_{i=1}^{m} (f(x_i) - y_i)^2 + \gamma \|f\|_K^2 \right\} \qquad (1)$$

## 3. Learning framework for improving the interpretability of the deep learning model

In traditional machine learning framework [11], generalization error bound[12] of prediction model is focused. However, if trying to shrink the gap between the deep learning model and cognitive model, in the learning process, not only the generalization error bound should be considered also the deviation boundary between the both models.

### 3.1 Evaluation of the interpretability of the prediction model

The cognitive model $P(x)$ firstly should be provided in a mathematical model describing prior knowledge. However, the prior knowledge is usually uncertain and incomplete. The uncertain representation method of the prior knowledge and how to obtain complete knowledge from incomplete knowledge, in another article [13], has been introduced.

We posit that the correctness of the deep learning model $f(x)$ itself depends on the correct expression of causal relationship. When the attributes in both models satisfy the same causal relationship, we can assume that both models express the same interpretation, even if their magnitude is different. Inspired by induction and analysis coupling learning method, in the square integral function space $\mathcal{L}_\rho^2(X)$ the variance $\mathcal{E}^P(f)$ of the error between the both model is used to calculate the gap which can evaluated the interpretability of $f(x)$.

$$\mathcal{E}^P(f) = \int_Z \left( f(x) - P(x) - \mu^P(f) \right)^2 \qquad (2)$$

where $\mu^P(f) = \int_Z (f(x) - P(x))$ is a mean error between $f(x)$ and $P(x)$.

## 3.2 Learning framework for improving the interpretability of the deep learning model

Based on Tihonov regularized learning framework [10-11] and the evaluation formula of the interpretability, a learning framework for improving the interpretability of the deep learning model can be obtained.

$$f_{z,\lambda} = \operatorname{argmin}_{f \in \mathcal{H}_K} \left\{ \frac{1}{m}\sum_{i=1}^{m}(f(x_i) - y_i)^2 + \lambda \|f\|_K^2 + \frac{1}{m}\sum_{i=1}^{m}\left(f(x_i) - P(x_i) - \frac{1}{m}\sum_{i=1}^{m}|f(x_i) - P(x_i)|\right)^2 \right\} \qquad (3)$$

## 4. Error estimate of Hypothesis Space

Suppose the optimal solution $f_{\mathcal{H}}^P(x)$ of the optimal problem (6) can be found in the convex subset $\mathcal{H}$ of $\mathcal{L}_\rho^2(X)$. The deviation between $f \in \mathcal{H}$ and $f_{\mathcal{H}}^P(x)$ is defined as an error $\mathcal{E}_\mathcal{H}(f) = \mathcal{E}(f) - \mathcal{E}(f_{\mathcal{H}}^P) + \mathcal{E}^P(f) - \mathcal{E}^P(f_{\mathcal{H}}^P)$, where $\mathcal{E}(f)$ is an error between $f(x)$ and the real output $y$. If $f: X \to Y$, $\mathcal{E}(f) = \mathcal{E}_\rho(f) = \int_Z (f(x) - y)^2$.

For any function $f \in \mathcal{H}$, $\mathcal{E}_\mathcal{H}(f) \geq 0$ and $\mathcal{E}_\mathcal{H}(f_{\mathcal{H}}^P) = 0$. Let us focus on that

$$\mathcal{E}(f_{\mu z}) + \mathcal{E}^P(f_{\mu z}) = \mathcal{E}_\mathcal{H}(f_{\mu z}) + \mathcal{E}(f_{\mathcal{H}}^P) + \mathcal{E}^P(f_{\mathcal{H}}^P) \qquad (4)$$

where $\mathcal{E}_\mathcal{H}(f_{\mu z})$ is a distance between $f_{\mu z}(x)$ and $f_{\mathcal{H}}^P(x)$, denoted by sample error. $\mathcal{E}(f_{\mathcal{H}}^P)$ is a distance between $f_{\mathcal{H}}^P(x)$ and $y$, and $\mathcal{E}^P(f_{\mathcal{H}}^P)$ is a distance between $f_{\mathcal{H}}^P(x)$ and $P(x)$, the sum of the two distances is approximate error.

### 4.1 Sample error estimation

From the above formula(4), it can be seen that

$$\mathcal{E}_\mathcal{H}(f) = \mathcal{E}(f) - \mathcal{E}(f_{\mathcal{H}}^P) + \mathcal{E}^P(f) - \mathcal{E}^P(f_{\mathcal{H}}^P) \qquad (5)$$

The formula can be divided into two parts: $\mathcal{E}(f) - \mathcal{E}(f_{\mathcal{H}}^P)$ and $\mathcal{E}^P(f) - \mathcal{E}^P(f_{\mathcal{H}}^P)$.

Probability bound of the former, $\mathcal{E}(f) - \mathcal{E}(f_{\mathcal{H}}^P)$, can be deduced by Theorem B, theorem C and lemma 5 in literature [14], it is easy to deduce the following conclusion.

**Theorem 1**. Suppose $\mathcal{H}$ is a compact convex subset of $\mathcal{L}_\rho^2(X)$. If to all $f \in \mathcal{H}$, $P(x)$ is an interpretation function, $|f(x_i) - P(x_i) - \mu^P(f)| \leq M_P$, and $|f(x) - y| \leq M$ is true almost everywhere, for all $\varepsilon_P > 0$, inequality

$$prob_{z \in Z^m}\{|\mathcal{E}^P(f_z) - \mathcal{E}^P(f_{\mathcal{H}}^P)| + |\mathcal{E}(f_z) - \mathcal{E}(f_{\mathcal{H}}^P)| \leq \varepsilon\} \geq$$

$$1 - \mathcal{N}\left(\mathcal{H}, \frac{\varepsilon}{8(3M + 2M_P)}\right) e^{-\frac{m\varepsilon}{32(M^2 + M_P^2)}\left(\frac{M}{3M + 2M_P}\right)^2} \quad (6)$$

holds.

## 4.2 Approximation error estimate

Based on the Hilbert-Schmidt theorem, we can get the theorem 2.

**Theorem 2.** Suppose $\mathcal{H}$ is a Hilbert space, $A$ is a strict positive definite self adjoint compact operator.

(1) If $0 < r \leq s, r \in \mathbb{R}$, for all $a \in \mathcal{H}$, let $\mathcal{L} = Id - \Gamma$, $\Gamma(b - p) = \int (b - p) d\rho$, then

$$\min_{b \in \mathcal{H}}(\|b - a\|^2 + \tau\|b - p - \int(b-p)d\rho\|^2 + \gamma\|A^{-s}b\|^2) \leq$$
$$\|(Id + \tau\mathcal{L}^2 + \gamma A^{-2s})^{-1}[\tau\mathcal{L}^2 p - (\tau\mathcal{L}^2 + \gamma A^{-2s})a]\|^2 + \tau\|\mathcal{L}(Id + \tau\mathcal{L}^2 + \gamma A^{-2s})^{-1}[a -$$
$$(1 + \gamma A^{-2s})p]\|^2 + (r + s)^{\frac{r+s}{s}} \gamma^{\frac{r}{s}} (s - r)^{-\frac{r+s}{s}} (1 + \tau\mathcal{L}^2)^{-\frac{r+s}{s}} \|A^{-r}(a + \tau\mathcal{L}^2 p)\|^2 \quad (7)$$

(2) If $\|A^{-s}b\| \leq R, R > 0$, for all $a \in \mathcal{H}$,

$$\min_{b \in \mathcal{H}}(\|b - a\|^2 + \tau\|b - p - \int(b-p)d\rho\|^2) \leq \|(Id + \tau\mathcal{L}^2 + \gamma A^{-2s})^{-1}[\tau\mathcal{L}^2 p -$$
$$(\tau\mathcal{L}^2 + \gamma A^{-2s})a]\|^2 + \tau\|\mathcal{L}(Id + \tau\mathcal{L}^2 + \gamma A^{-2s})^{-1}[a - (1 + \gamma A^{-2s})p]\|^2 \quad (8)$$

where $\gamma \leq (r + s)^{\frac{r+s}{s-r}} R^{-\frac{2s}{s-r}} (s - r)^{-\frac{r+s}{s-r}} (1 + \tau\mathcal{L}^2)^{-\frac{r+s}{s-r}} \|A^{-r}(a + \tau\mathcal{L}^2 p)\|^{\frac{2s}{s-r}}$.

In both cases, $b$ is uniquely exists and finite and in the first part, the optimal $b$ is

$$\hat{b} = (Id + \tau\mathcal{L}^2 + \gamma A^{-2s})^{-1}(a + \tau\mathcal{L}^2 p).$$

Now, in a Hilbert space, let us introduce a general setting. Suppose $\nu$ is a Borel measure in $X$ and $A: \mathcal{L}_\nu^2(X) \to \mathcal{L}_\nu^2(X)$ is a strict positive definite compact operator, and $\mathbb{E} = \{g \in \mathcal{L}_\nu^2(X) | \|A^{-s}g\|_\nu < \infty\}$ where $\mathcal{L}_\mu^2(X)$ is a squared integrable function space with Lebesgue measure $\mu$ induced by $\mathbb{R}^n$ quotient space $X$. In $\mathbb{E}$, an inner product is defined as $\langle g, h \rangle_{\mathbb{E}} = \langle A^{-s}g, A^{-s}h \rangle_\nu$. $\mathbb{E}$ is a Hilbert space. So, $A^{-s}: \mathcal{L}_\nu^2(X) \to \mathbb{E}$ is a Hilbert isomorphism. For the general setting, some supposes should be given. $\mathbb{E} \to \mathcal{L}_\nu^2(X)$ can be decomposed into $J_{\mathbb{E}}: \mathbb{E} \to C(X)$ and $C(X) \subset \mathcal{L}_\nu^2(X)$. Suppose $\mathcal{H} = \mathcal{H}_{\mathbb{E},R}$ is $\overline{J_{\mathbb{E}}(B_R)}$, where $B_R$ is a sphere with radius $R$. If $\mathcal{D}_{\nu\rho}$ is a norm of operator $J: \mathcal{L}_\nu^2(X) \to \mathcal{L}_\rho^2(X)$, we can obtain Theorem 3.

**Theorem 3.** In the general setting of a Hilbert space, for $0 < r \leq s, r \in \mathbb{R}$, the approximation error

$$\mathcal{E}(f_{\mathcal{H}}^P) + \mathcal{E}^P(f_{\mathcal{H}}^P) = \min_{g(x) \in B_R} \left( \|f_\rho(x) - g(x)\|_\rho^2 + \tau\|g(x) - P(x) - \mu^P(g)\|_\rho^2 \right) + \sigma_\rho^2 \leq$$

$$\mathcal{D}_{\nu\rho}^2 \|(Id + \tau\mathcal{L}^2 + \gamma A^{-2s})^{-1}[\tau\mathcal{L}^2 P(x) - (\tau\mathcal{L}^2 + \gamma A^{-2s})f_\rho(x)]\|_\nu^2 + \tau\mathcal{D}_{\nu\rho}^2 \|\mathcal{L}(Id + \tau\mathcal{L}^2 +$$

$$\gamma A^{-2s})^{-1}[f_\rho(x) - (1 + \gamma A^{-2s})P(x)]\|_\nu^2 + \sigma_\rho^2 \tag{9}$$

where

$$\gamma \leq (r+s)^{\frac{r+s}{s-r}} R^{-\frac{2s}{s-r}} (s-r)^{-\frac{r+s}{s-r}} (1+\tau\mathcal{L}^2)^{-\frac{r+s}{s-r}} \mathcal{D}_{\nu\rho}^2 \left\| A^{-r}\left(f_\rho(x) + \tau\mathcal{L}^2 P(x)\right)\right\|_\nu^{\frac{2s}{s-r}}.$$

If $\nu = \rho$, $\mathcal{D}_{\nu\rho} = 1$.

### 4.3 Approximation error estimate in Sobolev space

In the section, suppose $X \subset \mathbb{R}^n$ is a compact region with smooth boundary.

**Theorem 4**. If $0 < r < s$, $B_R$ is a sphere with radius $R$ in a conjugate space $H(X)$ on $X$, and $\mathcal{H} = \overline{J_{H(X)}(B_R)}$, the approximation error is

$$(f_{\mathcal{H}}^P) + \mathcal{E}^P(f_{\mathcal{H}}^P) \leq \mathcal{D}_{\nu\rho}^2 \|(Id + \tau\mathcal{L}^2 + \gamma A^{-2s})^{-1}[\tau\mathcal{L}^2 P(x) - (\tau\mathcal{L}^2 + \gamma A^{-2s})f_\rho(x)]\|_\nu^2 +$$

$$\tau\mathcal{D}_{\nu\rho}^2 \|\mathcal{L}(Id + \tau\mathcal{L}^2 + \gamma A^{-2s})^{-1}[f_\rho(x) - (1 + \gamma A^{-2s})P(x)]\|_\nu^2 + \sigma_\rho^2 \tag{10}$$

where $\gamma \leq (r+s)^{\frac{r+s}{s-r}} (RC)^{-\frac{2s}{s-r}} (s-r)^{-\frac{r+s}{s-r}} (1+\tau\mathcal{L}^2)^{-\frac{r+s}{s-r}} \mathcal{D}_{\nu\rho}^2 \|f_\rho(x) + \tau\mathcal{L}^2 P(x)\|_\nu^{\frac{2s}{s-r}}$,

and $C$ is a constant only depends on $r, X$.

## 5. How to solve this new learning problem

According the general setting in section 4.2, suppose sample size is $m$ and the confidence is $1 - \delta, 0 < \delta < 1$. For every $R > 0$, hypothesis space $\mathcal{H} = \mathcal{H}_{\mathbb{E},R}$. We consider $f_{\mathcal{H}}^P$ and $f_z, z \in Z^m$. In the general setting, the optimal solution of the new learning problem (6) is how to find an optimal $R$, $M$ and $M_P$, where $|f(x) - P(x) - \mu^P(f)| \leq M_P$, and $|f(x) - y| \leq M$ are true almost everywhere.

**Theorem 5**. For all $m \in \mathbb{N}$, $\delta \in \mathbb{R}, 0 < \delta < 1$, and $r \in \mathbb{R}, 0 < r < s$, in the general setting, the optimal $R^*$, $M^*$ and $M_P^*$, can be found in the learning framework for improving the interpretability of a predication model.

The optimal $R^*$, $M^*$ and $M_P^*$, can be found in the learning framework for improving the interpretability of the predication model. And $\|J_\mathbb{E}\|R^* + M_\rho + \|f_\rho\|_\infty = M^*$ and $M^* = \frac{-(M_\rho+\|f_\rho\|_\infty)+\sqrt{(M_\rho+\|f_\rho\|_\infty)^2-24M_P^{*2}}}{4}$ will minimize the sample error and the approximate error.

## 6. Conclusion

In this paper, we proposed a quantitative index of the interpretability, and analyzed the relationship between the interpretability and the generalization performance of the deep learning model. The equilibrium problem between the two performances was proven to exist. We studied a universal learning framework for improving the interpretability of the deep learning model. Next, the uniqueness of solution of the problem was proved and condition of unique solution was found. Probability upper bound of the sum of the two performances is analyzed. The solving method was proposed for the equilibrium problem.

## Acknowledgments

We would like to acknowledge support for this project from the National Natural Science Foundation of China (Program Numbers 61672027, 61773314, 61773313), the National Key R&D Program of China(Program Number 2017YFB1201500), the Natural Science Foundation of Shaanxi Province (Program Numbers 2014JQ8299, 2017JM6080) and the Scientific Research Program Funded by ShaanxiProvincial Education Department (Program Number 18JS076).